\documentclass[10pt, a4paper]{article}

\usepackage[]{lrec2026}
\usepackage{CJKutf8}
\usepackage{enumitem}
\usepackage{tabularx, booktabs} 
\usepackage{float}
\usepackage{graphicx}
\usepackage{comment}
\usepackage{multirow}  
\usepackage{xcolor}
\usepackage{soul}
\definecolor{loancolor}{RGB}{255,230,230}
\definecolor{nativecolor}{RGB}{230,255,230}
\usepackage{tcolorbox}
\tcbuselibrary{listings,skins}
\usepackage{makecell}
\usepackage{rotating}
 
\usepackage{listings}
\lstdefinestyle{promptstyle}{
  basicstyle=\ttfamily\small,
  frame=single,
  breaklines=true,
  keepspaces=true,
  captionpos=b
}

\title{Are Language Models Borrowing-Blind? A Multilingual Evaluation of Loanword Identification across 10 Languages}

\name{Mérilin Sousa Silva \quad\quad\quad\quad\quad Sina Ahmadi} 

\address{Department of Computational Linguistics \\
         University of Zurich \\
         \{merilin.sousasilva,sina.ahmadi\}@uzh.ch\\}

\abstract{
Throughout language history, words are borrowed from one language to another and gradually become integrated into the recipient's lexicon. Speakers can often differentiate these loanwords from native vocabulary, particularly in bilingual communities where a dominant language continuously imposes lexical items on a minority language. This paper investigates whether pretrained language models, including large language models, possess similar capabilities for loanword identification. We evaluate multiple models across 10 languages. Despite explicit instructions and contextual information, our results show that models perform poorly in distinguishing loanwords from native ones. These findings corroborate previous evidence that modern NLP systems exhibit a bias toward loanwords rather than native equivalents. Our work has implications for developing NLP tools for minority languages and supporting language preservation in communities under lexical pressure from dominant languages.\\ \newline \Keywords{lexical borrowing, multilingualism, low-resourced languages} }

\begin{document}

\maketitleabstract

\section{Introduction}

Loanwords or lexical borrowings are words adopted from one language into another through language contact. The English lexicon illustrates this phenomenon vividly: \textit{sugar}, \textit{cushy}, \textit{sky}, and \textit{chocolate} all entered English at different historical periods, reflecting geopolitical, cultural, and commercial exchanges with other languages. Lexical borrowing rates vary greatly among languages, with average borrowing rate at 24.2\% based on the Loanword Typology (LWT) project~\cite{haspelmath2009loanword}. However, these rates also reflect varying degrees of knowledge about each language's history. Despite such measurement challenges, it is clear that borrowing occurs in all languages as every language contains at least some words adopted from contact with other linguistic communities \cite{tadmor2009loanwords}.

The impact of lexical borrowing extends beyond individual word histories. In bilingual and multilingual communities, the flow of loanwords from dominant to minority languages can accelerate language shift and vocabulary loss \cite{rabiu2024borrowing,sokpo2020lexical}. Research has identified two primary drivers of borrowing: need, when languages adopt terms for new concepts, and prestige, when speakers borrow from socially dominant languages even for existing concepts. While all languages borrow regardless of their status, prestige predicts the direction of borrowing as lexical items flow from more powerful to less powerful linguistic communities, with this asymmetry intensifying during periods of significant cultural contact~\citet{carling2019causality}. Understanding and identifying loanwords is therefore crucial not only for historical linguistics but also for language preservation efforts, educational initiatives, and developing tools in computational linguistics (CL) and natural language processing (NLP).

Although borrowing has been long studied in linguistics, as in \cite{poplack1988social,winford2010contact,van1994modeling} to mention but a few, it has received limited attention in CL and NLP where previous studies focus on a handful of languages and are restricted to narrow aspects of lexical borrowing, such as loanword detection based on word lists \cite{DBLP:journals/talip/MiXZ20,DBLP:conf/eacl/MillerL23} or analysis of anglicisms in specific languages \cite{alvarez2025overview,DBLP:conf/acl-codeswitch/Alvarez-Mellado20}. To further clarify the computational challenges of lexical borrowing, recent work introduced ConLoan \cite{ahmadi2025conloan}, a contrastive multilingual dataset for evaluating how NLP systems handle loanwords versus native alternatives. That work demonstrated that neural machine translation (NMT) systems and large language models (LLMs) show systematic preferences for loanwords in generation and comprehension tasks, exhibiting higher surprisal for native alternatives and reduced translation quality when loanwords are replaced. Building on these findings, the present study makes two-fold contributions:

\begin{enumerate}[noitemsep]
\item We directly investigate whether models can identify loanwords when explicitly asked to do so, evaluating various pretrained models on loanword identification across 10 typologically diverse languages using the ConLoan dataset. Our findings show that general-purpose LLMs consistently fail to distinguish borrowed from native vocabulary.
\item We fine-tune several pretrained models and demonstrate, through both quantitative and qualitative analysis, how different architectures and training approaches perform on this task. Our experiments indicate that loanword identification remains far from solved, even with task-specific fine-tuning.
\end{enumerate}

\section{Related Work}

Loanwords are lexical items borrowed from one language and integrated into another language's vocabulary, where they become part of the recipient language's lexicon and are used by its speakers, including monolinguals. Identifying loanwords has proven beneficial across multiple domains, including language contact~\cite{gardani2022contact}, language education~\cite{dashti2017morphological}, language technology~\cite{DBLP:journals/talip/Mi23}, and speech recognition~\cite{paik2025hikehierarchicalevaluationframework,hirai2024speech}. For language preservation efforts, accurate loanword identification helps document lexical influence patterns and supports native vocabulary development, while in educational contexts, it aids etymology instruction and language history teaching. For NLP systems, understanding lexical borrowing is essential for developing multilingual models that can properly handle diverse linguistic phenomena. The task of loanword identification is closely related to code-switching detection, which has been extensively studied in NLP~\cite{bali-etal-2014-borrowing, DBLP:conf/naacl/JamesYS0KMJ22}. Both phenomena involve mixing linguistic material from multiple languages, but they differ fundamentally: code-switching involves alternating between languages within discourse, while borrowing represents lexical adoption into a recipient language's system \cite{alvarez-mellado-lignos-2022-borrowing}.

Previous computational approaches to loanword identification have employed diverse feature sets and methodologies. Phonetic, phonological and morphosyntactic features have been widely used, leveraging the observation that loanwords often retain patterns from donor languages~\cite{DBLP:journals/corr/abs-2508-17923}. \citet{patro-etal-2017-english} introduced methods to identify borrowing likelihood based on social media content, demonstrating that lexical innovation and borrowing patterns are observable in online language use. More recent work has incorporated semantic similarity metrics from embedding models~\cite{serigos2022using} and LLMs \cite{mi2025loanword}. Much of this work has focused on specific language pairs, particularly anglicisms in various recipient languages~\cite{DBLP:conf/coling/NathSKMLK22}. Nevertheless, loanword identification remains a challenging and largely unsolved task, with significant room for improvement particularly for low-resource languages and contexts where borrowed terms are highly integrated~\cite{DBLP:journals/cin/MiZN21}.

Recent work on ConLoan~\cite{ahmadi2025conloan}, a contrastive multilingual dataset where loanwords are paired with native alternatives across 10 languages, show that modern NLP systems exhibit systematic biases toward loanwords: LLMs (EuroLLM, Llama 2.7 and 3.1) show lower surprisal for sentences containing loanwords compared to those with native alternatives, and NMT models (NLLB) perform less efficiently when translating sentences with native replacements. These findings raise a critical question: \textbf{If models prefer loanwords in generation and comprehension tasks, can they identify loanwords when explicitly instructed to do so?} Existing work has not systematically evaluated whether general-purpose pretrained models and LLMs possess this capability, nor explored whether fine-tuning can enable them to distinguish borrowed from native vocabulary. This paper fills this gap by directly testing models' ability to perform loanword identification across multiple languages and training paradigms.

\section{Methodology}

We evaluate loanword identification capabilities through two complementary approaches: (1) prompting large language models in zero-shot and few-shot (two-shot) settings, and (2) fine-tuning multilingual token classification models. Both approaches use the ConLoan dataset~\cite{ahmadi2025conloan}, which provides sentence-level annotations with loanword spans across Chinese, French, German, Greek, Icelandic, Italian, Northern Kurdish, Portuguese, Russian, and Spanish. The following is an example in Portuguese where the loanwords `\textit{franchise}` is replaced by a native alternative `\textit{franquia}` in the annotation process:

\begin{center}
\small
\parbox{0.95\columnwidth}{
\textit{Podemos vender-te um \colorbox{loancolor}{\textbf{franchise}} disto por 3000\$.} \\[0.2em]
{\textcolor{gray}{(We can sell you a franchise of this for 3,000 dollars.})} \\[0.2em]
\textit{Podemos vender-te um \colorbox{nativecolor}{\textit{\textbf{franquia}}} disto por 3000\$.} \\
}
\end{center}

\subsection{Task Formulation}

We formulate loanword identification as a sequence labeling task using Inside–outside–beginning (BIO) tagging (O, B-LOAN, I-LOAN). Given a sentence tokenized into words $w_1, w_2, \ldots, w_n$, the task is to assign each word a tag indicating whether it is outside a loanword span (O), begins a loanword span (B-LOAN), or continues a loanword span (I-LOAN). This formulation handles both single-word and multi-word loanwords uniformly. 

\subsection{LLMs}

We evaluate three LLMs: Gemini-2.5-Flash-Lite, GPT-4.1, and Meta-Llama-3-8B-Instruct. Each model is tested under two setups:

\begin{itemize}[noitemsep,leftmargin=*]
    \item \textbf{Zero-shot:} Models receive only task instructions without examples.
    \item \textbf{Few-shot:} Models are provided with the same two in-language exemplars of the target language before processing the test sentence.
\end{itemize}

How loanwords are defined determines how they are identified and this makes this task particularly compelling. To assess how the internal knowledge of the selected LLMs might (or not) affect the identification task, we evaluate three prompt variants that progressively add more granularity to the definition of loanwords:

\begin{itemize}[noitemsep]
\item \textbf{Prompt 1 (Minimal):} Basic instruction to detect loanwords without definition.

    \begin{tcolorbox}[
                colback=gray!5!white,
                colframe=gray!75!black,
                boxrule=0.5pt,
                arc=1pt,
                fontupper=\ttfamily\footnotesize
            ]
            You are a loanword detection system. Identify loanwords in: "\{sentence\}"
            \end{tcolorbox}

\item \textbf{Prompt 2 (Etymological):} Adds \citet{haspelmath2009loanword}'s definition emphasizing historical borrowing.

    \begin{tcolorbox}[
            colback=gray!5!white,
            colframe=gray!75!black,
            boxrule=0.5pt,
            arc=1pt,
            fontupper=\ttfamily\footnotesize
        ]
    You are a loanword detection system. Loanword (or lexical borrowing) is here defined as a word that at some point in the history of a language entered its lexicon as a result of borrowing (or transfer, or copying). Identify loanwords in: "\{sentence\}"
    \end{tcolorbox}
    
    \item \textbf{Prompt 3 (Usage-based):} Extends \citet{haspelmath2009loanword}'s definition by focusing on conventional use by monolinguals, distinguishing loanwords from code-switching.

    \begin{tcolorbox}[
        colback=gray!5!white,
        colframe=gray!75!black,
        boxrule=0.5pt,
        arc=1pt,
        fontupper=\ttfamily\footnotesize
        ]
    You are a loanword detection system. From the point of view of an entire language (not that of a single speaker), a loanword is a word that can conventionally be used as part of the language. In particular, it can be used in situations where no code-switching occurs, e.g. in the speech of monolinguals. This is the simplest and most reliable criterion for distinguishing loanwords from single-word switches. Identify loanwords in: "\{sentence\}"
    \end{tcolorbox}
\end{itemize}

All prompts constrain output to valid Python lists of strings to enable deterministic parsing. In addition, regular expressions are employed as a supplementary method to identify lists within the LLM's responses.
These responses are normalized and cached using MD5 hashing to ensure reproducibility and avoid redundant API calls.

\subsection{Multilingual Encoder Models}

We evaluate four pretrained multilingual encoders: mBERT\footnote{\texttt{bert-base-multilingual-cased}}, base and large variants of XLM-RoBERTa\footnote{\texttt{xlm-roberta-base/large}}, and ELECTRA-base multilingual\footnote{\texttt{google/electra-base-discriminator}}. Each model is tested in two configurations:

\begin{itemize}[noitemsep]
    \item \textbf{Zero-shot baseline:} Models are used as frozen feature extractors without task-specific training. A deterministic rule based on subword continuity and token boundaries maps hidden states to BIO tags.
    \item \textbf{Fine-tuned:} Models are fine-tuned with a token classification head on an 80/20 train/test split of ConLoan. We use the following hyperparameters: 10 epochs, batch size 16, AdamW optimizer with weight decay 0.01. Training employs cross-entropy loss with label -100 for special tokens and subsequent subword pieces to exclude them from gradient computation.
\end{itemize}

\subsection{Evaluation Metrics}

We compute precision, recall, and F1-score using the seqeval library\footnote{\url{https://github.com/chakki-works/seqeval}}, which evaluates at the entity span level rather than per-token. This penalizes boundary errors appropriately for the identification task. We report both overall (macro-averaged across languages) and per-language metrics. For LLM evaluation, we employ two scoring protocols:

\begin{itemize}[noitemsep,leftmargin=*]
    \item \textbf{Strict:} Set-based comparison where predicted spans must match gold segmentation exactly. Multi-word loanwords must appear as single items if annotated as such.
    \item \textbf{Relaxed:} Gold multi-word spans are decomposed into their constituent tokens prior to comparison, ensuring that both \texttt{['social media']} and \texttt{['social', 'media']} are considered equivalent matches. This procedure was adopted because LLMs exhibited difficulty in preserving multi-word loanwords as single tokens. Consequently, this approach accommodates segmentation variability while maintaining strict requirements for correct token coverage. 
\end{itemize}

\section{Experiments}

\subsection{LLMs Performance}

\begin{table*}[t]
\centering
\begin{tabular}{lrrr|rrr}
\toprule
\multirow{2}{*}{Language} & \multicolumn{3}{c}{Prompt-based} & \multicolumn{3}{c}{Model-based} \\ \cline{2-7}
  & Prompt1 & Prompt2 & Prompt3  & Gemini & Llama & OpenAI \\
\midrule
Chinese & 0.605 & 0.521 & 0.537 & 0.697 & 0.429 & 0.537 \\
French & 0.474 & 0.361 & 0.453 & 0.557 & 0.254 & 0.476 \\
German & 0.141 & 0.129 & 0.120 & 0.168 & 0.086 & 0.135 \\
Greek & 0.391 & 0.335 & 0.327 & 0.513 & 0.178 & 0.362 \\
Icelandic & 0.307 & 0.285 & 0.299 & 0.382 & 0.149 & 0.359 \\
Italian & 0.469 & 0.384 & 0.440 & 0.516 & 0.259 & 0.517 \\
Northern-Kurdish & 0.438 & 0.392 & 0.378 & 0.489 & 0.200 & 0.519 \\
Portuguese & 0.325 & 0.260 & 0.324 & 0.350 & 0.261 & 0.298 \\
Russian & 0.397 & 0.387 & 0.412 & 0.502 & 0.201 & 0.493 \\
Spanish & 0.432 & 0.314 & 0.436 & 0.486 & 0.293 & 0.402 \\\hline
Average & \textbf{0.3979} & 0.3368 & 0.3726 & \textbf{0.466} & 0.231 & 0.4098 \\
\bottomrule
\end{tabular}
\caption{F1 scores per language for each model and prompt (averaged over all prompts and configurations)  in loanword identification using LLMs. Not providing explicit definition (Prompt 1) and using Gemini result in higher F1.}
\label{tab:avg_by_prompt_lang}
\end{table*}

\begin{figure}[b]
\centering
\includegraphics[width=\linewidth]{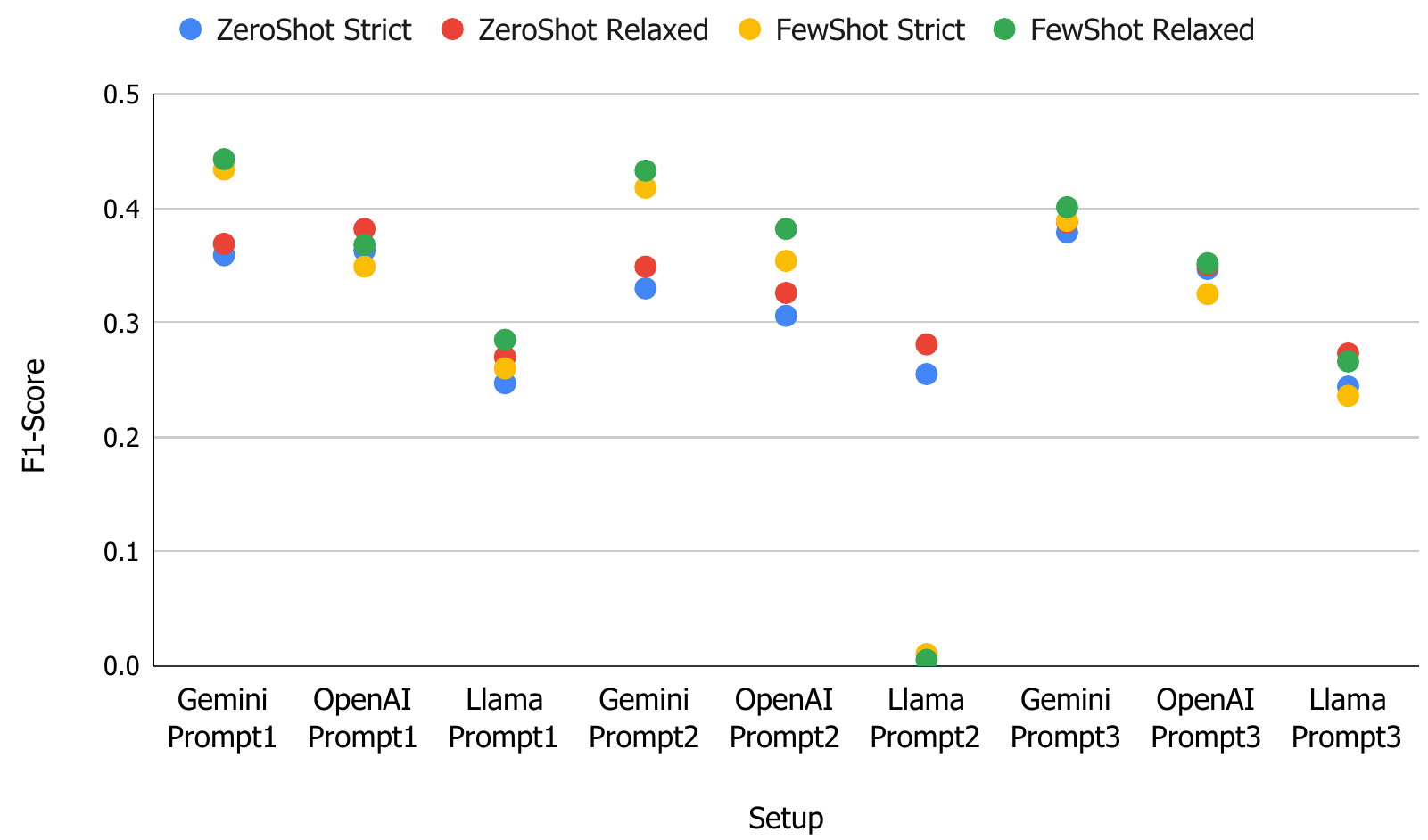}
\caption{F1-scores of LLMs based on prompt, evaluation protocol, and fine-tuning setup. Overall, providing few shots with relaxed evaluation using Gemini yields higher F1-score for loanword identification.}
\label{fig:OverallLLM}
\end{figure}

Table \ref{tab:avg_by_prompt_lang} presents the performance of each model and prompt configuration. In general, the results reveal that the detection of loanwords remains a difficult task for LLMs; in all settings, the F1 scores stay below 0.70. Among the models tested, Gemini achieved the best overall performance (0.466), followed by OpenAI (0.4098) and Llama (0.231). Adding few-shot examples had mixed effects. In some cases, as in Prompts 1 and 2, it slightly improved results (by 0.01–0.09 points) by helping models achieve higher recall. However, for Prompt 3, performance actually declined (by 0.02–0.04 points), as shown in Figure~\ref{fig:OverallLLM}.

A particularly striking anomaly appeared with Llama under the second few-shot prompt, where F1-scores nearly dropped to zero, which is far below its usual average of about 0.35. We suspect this happened because the added examples shifted the model’s internal expectations, causing it to ignore the original instruction format. Instead of returning structured lists as required, Llama began producing free-form text or syntactically broken lists. Since our evaluation relied on strict parsing of list outputs, these deviations were counted as empty predictions, collapsing the F1-score even though the underlying reasoning capability may not have degraded as drastically.

\textbf{Language Performance:} Performance varies notably across languages. Chinese, French and Italian achieve the highest average F1-scores across both prompt-based and model-based settings, suggesting that loanword boundaries in these languages are comparatively easier for models to identify. In contrast, German, Icelandic, and Portuguese yield the lowest results, likely due to their rich morphological inflection, compounding tendencies, and the greater integration of loanwords into native morphology. Similarly, it can be seen that among the model-based systems, Gemini consistently achieves the strongest results across most languages, followed by OpenAI and Llama.

\begin{table*}
\centering
\setlength{\tabcolsep}{2.5pt}
\begin{tabular}{lrrrr|rrrr} 
\toprule
\multirow{2}{*}{Language} & \multicolumn{4}{c}{Zero-shot} & \multicolumn{4}{c}{Fine-tuned} \\ 
\cline{2-9}
 & \multicolumn{1}{l}{ELECTRA} & \multicolumn{1}{l}{mBERT} & \multicolumn{1}{l}{XLM-R\textsubscript{B}} & \multicolumn{1}{l}{XLM-R\textsubscript{L}} & \multicolumn{1}{|l}{ELECTRA} & \multicolumn{1}{l}{mBERT} & \multicolumn{1}{l}{XLM-R\textsubscript{B}} & \multicolumn{1}{l}{XLM-R\textsubscript{L}}  \\ \hline
Northern Kurdish & 0.034 & 0.034 & 0 & 0.028 & 0.622 & 0.683 & 0.705 & 0.735 \\
Russian & 0.031 & 0.046 & 0 & 0.057 & 0.504 & 0.691 & 0.778 & 0.829 \\
Italian & 0.022 & 0.031 & 0.009 & 0.015 & 0.831 & 0.888 & 0.921 & 0.934 \\
Portuguese & 0.015 & 0.015 & 0.003 & 0.019 & 0.676 & 0.731 & 0.755 & 0.775 \\
French & 0.014 & 0.019 & 0 & 0.016 & 0.924 & 0.942 & 0.913 & 0.935 \\
Spanish & 0.011 & 0.015 & 0 & 0.016 & 0.816 & 0.914 & 0.898 & 0.917 \\
German & 0.01 & 0.018 & 0 & 0.008 & 0.688 & 0.834 & 0.856 & 0.881 \\
Chinese & 0.009 & 0.01 & 0 & 0.004 & 0.244 & 0.775 & 0.84 & 0.815 \\
Greek & 0.005 & 0.005 & 0 & 0 & 0.423 & 0.783 & 0.827 & 0.848 \\
Icelandic & 0.003 & 0.004 & 0 & 0.003 & 0.752 & 0.818 & 0.829 & 0.844 \\ \hline
Average & 0.0154 & 0.0197 & 0.0012 & 0.0166 & 0.648 & 0.8059 & 0.8322 & \textbf{0.8513} \\
\bottomrule
\end{tabular}
\label{tab:MultilingLangDep}
\caption{F1-scores per language and model in two setups of zero-shot and finetuned. Subscript B and L refer to base and large variants, respectively. Fine-tuned XML-R achieves the highest F-score in loanword identification while the pretained unfine-tuned models perform poorly on the task. These models also outperform LLMs.}
\end{table*}

\subsection{Multilingual Encoder Performance}

Overall, fine-tuned multilingual encoders substantially outperform their unfine-tuned counterparts across all evaluation metrics and languages. As shown in Table 2, untrained models perform poorly, with average F1-scores below 0.2 across architectures and a maximum of only 0.0197 for Russian under XLM-R\textsubscript{L}. In contrast, fine-tuning leads to drastic improvements, yielding average F1-scores between 0.648 and 0.851, depending on the model. Among fine-tuned models, XLM-R\textsubscript{L} achieves the strongest overall performance (0.8513). 
Our findings show that pretraining on multilingual corpora lays a necessary foundation but does not go far enough for nuanced challenges like loanword identification.

\textbf{Language Performance:} At the language level, fine-tuning particularly benefits languages that previously exhibited low zero-shot performance, such as Chinese, Icelandic, and Northern Kurdish, where F1-scores rise from near-zero to above 0.7–0.8. Romance languages (French, Italian, Spanish, Portuguese) consistently reach the highest scores, reaching or even exceeding 0.8 F1 in most fine-tuned configurations, while German and Greek achieve slightly lower but still strong results (around 0.85 F1 under XLM-R\textsubscript{L}).

\section{Analysis}

We conduct a qualitative analysis of misclassified instances to better understand the underlying reasons for the failures of models and to examine whether fine-tuned models continue to display systematic errors.

\subsection{Code-Switching vs. Loanwords}

Across the model predictions, both LLMs and fine-tuned models exhibited difficulties in accurately identifying words as non-loanwords, which we would classify as code-switches. Code-switching refers to the practice of alternating between two or more languages, dialects, or accents while speaking. However, code-switches and loanwords are distinct linguistic phenomena, a distinction that the models generally fail to capture. For instance, LLMs frequently misclassified code-switches as loanwords such as the word \textit{really} in the following sentence in Northern Kurdish:

\begin{center}
\begin{quote}
    \textit{\textbf{really} ha, nizanim, tiştek nabe tiştek tevlihev bûbe. Tu çi dikî tu hê li bajarê xwe yî an?} (\textcolor{gray}{\textit{Really}, huh, I don’t know, nothing’s happening, something is confusing. What are you doing; are you still in your city?})
\end{quote}
\end{center}

In this context, labeling `really' as a loanword is not correct, since `really' is not morphologically or phonologically integrated into Kurdish and is used purely as a code-switching element. However, this pattern did not persist throughout all LLM responses. When presented with Prompt 3, which explicitly distinguishes between loanwords and code-switches, classification precision improved. Nevertheless, certain LLMs, such as OpenAI’s models, tended toward the opposite extreme, labeling as code-switches words that merely resemble English orthography, even when these words are conventionally used in monolingual contexts. One such example is the word \textit{deal} in the following sentence in French:

\begin{center}
\begin{quote}
    \textit{Il nous appartient, dans la mesure du possible -- et je m'y emploie -- de faire en sorte que ce qui est globalement un bon \textbf{deal} entre les Américains et les Chinois, soit un aussi bon \textbf{deal} pour les Européens.} (\textcolor{gray}{It is up to us, insofar as possible, and I am working on it, to ensure that what is overall a good \textit{deal} between the Americans and the Chinese is also a good \textit{deal} for the Europeans.})
\end{quote}
\end{center}

Although the native alternative would be \textit{accord}, \textit{deal} is commonly used in French monolingual discourse and should therefore be classified as a loanword. Despite this, the models prompted under the third setup consistently categorized it as a non-loanword.

As such, both pretrained and fine-tuned models struggle to operationalize the nuanced boundary between code-switching and lexical borrowing. While fine-tuning improves recognition of overt language alternation, models still over-rely on orthographic cues and fail to account for sociolinguistic integration. This suggests that accurate detection of code-switching requires a deeper contextual understanding of speaker intent and linguistic assimilation—dimensions that current models are not yet equipped to capture effectively.

\subsection{Named Entities and Proper Nouns}

Across models, named entities and proper nouns consistently emerged as a major source for misclassifications. Both zero-shot and few-shot setups frequently mislabeled entities such as country names, organizations, and acronyms (e.g., \textit{NASA}, \textit{ESA}) as loanwords, resulting in a higher false-positive rate. This behavior likely stems from the orthographic or phonological resemblance of such tokens to foreign lexical patterns, which the models interpret as evidence of borrowings. Although fine-tuning led to a notable improvement in recall, these errors persisted. For instance, in the following sentence in German, the compound \textit{PISA-Studie} was incorrectly identified as a loanword despite being a named entity referring to the \textit{Programme for International Student Assessment}:

\begin{quote}
    \textit{An der Spitze der internationalen Rangliste laut der letzten \textbf{PISA-Studie} steht der Shanghai-Distrikt von China.} (\textcolor{gray}{At the top of the international rankings according to the latest \textit{PISA study} is China’s Shanghai district.})
\end{quote}

Additional evidence for this pattern is found in the multilingual predictions of XLM-RoBERTa, where tokens such as \textit{Jazz} in Italian or \textit{golfe} in Portuguese were erroneously tagged as loanwords, even though they occur as part of named entities or idiomatic expressions. Similarly, the model occasionally misclassified tokens like 
\begin{CJK*}{UTF8}{bkai}
飛船上
\end{CJK*}
(``on the spacecraft'') in Cantonese due to script-based unfamiliarity, which it mistakenly associated with foreign origin rather than contextual meaning. 

Few-shot prompting with instruction-tuned LLMs, such as LLaMA, reduced these errors in some cases but introduced new inconsistencies. For example, the LLaMA model frequently failed to distinguish between common nouns derived from proper names (e.g., \textit{rail} in French technical discourse) and genuine borrowings as in the following sentence:

\begin{quote}
    \textit{[…]l'examen des conseillers à la sécurité pour le transport par route, par \textbf{rail} ou par voie navigable de marchandises dangereuses.} (\textcolor{gray}{[…] the examination of safety advisers for the transport by road, by \textit{rail}, or by inland waterways of dangerous goods.})
\end{quote}

In such contexts, capitalization cues alone were insufficient for accurate disambiguation, suggesting that the models rely heavily on surface-level orthographic features rather than semantic or contextual grounding. Taken together, both pretrained and fine-tuned models exhibit a limited capacity to encode the pragmatic and discourse-level properties that distinguish named entities from lexical borrowings. 

\subsection{Scientific/Technical and Greco-Latin Vocabulary}

Another recurrent source of misclassifications involved scientific and technical terminology, particularly words derived from Greco-Latin roots. Such terms exhibit a complex linguistic status: while some are historically borrowed, many are fully assimilated and function as native lexical items within specialized domains. Models fine-tuned on the ConLoan dataset tended to under-detect these terms as loanwords, likely because they appear morphologically regular and orthographically integrated within their host languages. Conversely, LLMs in few-shot settings often exhibited over-detection, labeling scientific or Greco-Latin vocabulary such as \textit{filosofia} in Portuguese as loanwords purely based on etymological cues. In the following sentence in Icelandic, the term \textit{nítröt} (`nitrates') was classified as a native word by XLM-RoBERTa\textsubscript{L}:

\begin{quote}
    \textit{[…] þar sem \textbf{nítröt} geta breyst í \textbf{nítrít} og nítrósamín, og hvatti til þess að teknar yrðu upp góðar starfsvenjur í landbúnaði til þess að tryggja eins lágt nítratmagn og kostur er.} (\textcolor{gray}{[…] where \textit{nitrates} can turn into nitrites and nitrosamines, and encouraged the adoption of good agricultural practices to ensure the lowest possible nitrate levels.})
\end{quote}

Similarly, LLMs like Gemini frequently misclassified scientific expressions such as \textit{mazout} (French) or \textit{national} (Russian context) due to their shared morphological ancestry with Latin-derived words, despite their full lexical assimilation in monolingual usage. These findings highlight the tension between etymological origin and contemporary linguistic integration, an area where models lack clear conceptual grounding.

We believe that scientific and technical terms expose the models’ inability to reconcile historical borrowing with synchronic linguistic norms. Whereas fine-tuned models often fail to detect subtle loanword traces, general-purpose LLMs overgeneralize based on etymology, revealing a lack of sensitivity to contextual and disciplinary registers.

\vspace*{-0.2cm}
\section{Conclusion} 

This paper sheds light on loanword identification in 10 languages using pretrained and language models, widely used in modern NLP. This task has remained relatively underexplored largely because data distinguishing loanwords from native words are limited, even though such knowledge is pivotal for studying language contact and acquisition. Using ConLoan dataset, we evaluate the performance of a few LLMs and pretrained models for this task and show that, despite the differences in performance using different prompts, models and fine-tuning setups, the task is quite challenging for LLMs with an average F-score of less than 0.5. Fine-tuning pretrained models yields higher performance results with XLM-R (large) achieving 0.8513 F-score. While fine-tuning improves precision, both pretrained and instruction-tuned models remain prone to over-reliance on orthographic and etymological cues. Code-switches are often mistaken for borrowings, proper nouns for foreign insertions, and Greco-Latin scientific terms for recent loans. These systematic errors reveal that loanword detection is far from a solved problem.

\vspace*{-0.1cm}
\paragraph{Limitations} The core contribution of the paper lays in the additional empirical results that demonstrate that the studied LLMs in this paper have a proclivity to not detect or differentiate loanwords accurately. In other words, models do not behave as language purists. Future work should consider a more fine-grained identification task where loanwords are categorized based on their integration status or within a continuum~\cite{poplack1984borrowing}. Successful identification requires sensitivity to pragmatic context, speaker intent, and lexical assimilation, dimensions that go beyond surface form. Additionally, creating and analyzing models with controlled vocabulary where loanwords are automatically replaced by native alternatives should be explored. We will release the models upon the acceptance of the paper.

\vspace*{-0.1cm}
\paragraph{Ethics Statement} This research involves computational analysis of publicly available linguistic data and poses no ethical concerns. The ConLoan dataset used contains no personally identifiable information, and our experiments focus solely on linguistic classification tasks without generating new content that could be harmful or biased.

\section*{Acknowledgments}
Sina Ahmadi gratefully thanks the support of the UZH Postdoc Grant (reference number 269093).

\section{Bibliographical References}

\bibliographystyle{lrec2026-natbib}
\bibliography{references}

\end{document}